\documentclass[10pt,twocolumn,letterpaper]{article}
\pdfoutput=1
\usepackage{cvpr}
\usepackage{times}
\usepackage{epsfig}
\usepackage{graphicx}
\usepackage{amsmath}
\usepackage{amssymb}
\usepackage{booktabs}
\usepackage{hyperref}

\cvprfinalcopy 

\def\cvprPaperID{****} 

\begin{document}

\title{KeyDetect - Detection of anomalies and user based on Keystroke Dynamics}

\author{Soumyatattwa Kar\thanks{Department of Mechanical Engineering, Carnegie Mellon University, 5000 Forbes Ave, Pittsburgh, PA 15213}\\
{\tt\small soumyatk@alumni.cmu.edu}
\and
Abhishek Bamotra\footnotemark[1]   \thanks{Correspondence author 412-692-1807}\\
{\tt\small abamotra@alumni.cmu.edu}
\and
Bhavya Duvvuri\thanks{Department of Civil and Environmental Engineering, Carnegie Mellon University, 5000 Forbes Ave, Pittsburgh, PA 15213}\\
{\tt\small bduwuri@alumni.cmu.edu}
\and
Radhika Mohanan\footnotemark[1]\\
{\tt\small rmohanan@alumni.cmu.edu}
}

\maketitle

\begin{abstract}
Cyber attacks has always been of a great concern. Websites and services with poor security layers are the most vulnerable to such cyber attacks. The attackers can easily access sensitive data like credit card details and social security number from such vulnerable services. Currently to stop cyber attacks, various different methods are opted from using two-step verification methods like One-Time Password and push notification services to using high-end bio-metric devices like finger print reader and iris scanner are used as security layers. These current security measures carry a lot of cons and the worst is that user always need to carry the authentication device on them to access their data. To overcome this, we are proposing a technique of using keystroke dynamics (typing pattern) of a user to authenticate the genuine user. In the method, we are taking a data set of 51 users typing a password in 8 sessions done on alternate days to record mood fluctuations of the user. Developed and implemented anomaly-detection algorithm based on distance metrics and machine learning algorithms like Artificial Neural networks (ANN) and convolutional neural network (CNN) to classify the users. In ANN, we implemented multi-class classification using 1-D convolution as the data was correlated and multi-class classification with negative class which was used to classify anomaly based on all users put together. We were able to achieve an accuracy of 95.05\% using ANN with Negative Class. From the results achieved, we can say that the model works perfectly and can be bought into the market as a security layer and a good alternative to two-step verification using external devices. This technique will enable users to have two-step security layer without worrying about carry an authentication device.
\end{abstract}
\section{Introduction}
Cyber security has always been the most important concern about the internet usage. Everyday millions of people get attacked by impostors to extract their valuable information like credit card details, social security number etc. These information are vital to a person and hold a lot of value, exploitation of such information can result in great loss \cite{security}. Also, cyber security is a must and companies spend millions of dollar on improvement and constant development of security layers to protect sensitive data from getting attacked.

There are various type of security layers deployed in the internet sphere some of them are two-step verification using One-Time Password (OTP), push verification etc\cite{revett2007machine}. In such security layers, user is required to enter the credentials for the account, select two-step verification and wait for a message which will contain OTP or a push notification on their mobile. Using the message/notification, the user verifies the login. Such security layers are good in protecting the data even when an imposter is able to extract user's credentials. These layers are strong and helpful but requires a user to always carry their authentication device on them. So, loss or unavailability of the authentication device will result in blocking the user from accessing the data.

To tackle this issue, high-end security devices like finger print reader and iris scanners are used\cite{iris}. One proposed approach is using keystroke dynamics of a person to authenticate the user as second layer security\cite{originalpaper}\cite{idrus2014soft}. The technique uses the user's unique typing pattern and distinguish the genuine user from an imposter. Using keystroke dynamics, even if the imposter knows the credentials of a person, imposter won't be able to login as keystroke pattern of an imposter will vary significantly from the genuine user\cite{ali2017keystroke}\cite{muliono2018keystroke}. The dataset used in this problem is taken on a physical computer keyboard and can also be implemented on smartphones\cite{zahid2009keystroke}.

We have tested various anomaly-detection algorithms based on distance metrics and machine learning algorithms like Random forest, Support vector machine (SVM), Artificial Neural networks (ANN) and convolutional neural network (CNN) to detect genuine users (discussed in a later section). Accuracy or performance of an algorithm is the measure one uses to evaluate an algorithm, if the results from an algorithm makes sense or are just random outputs with minimal or no sense. To evaluate our distance metrics algorithms, we are using European standard norm of false-alarm rate and miss rate\cite{patil2016keystroke}. Whereas for machine learning algorithms, we are giving accuracy of the classifiers to evaluate the algorithms.

The results demonstrated by the CNN 1-D convolution and ANN with negative class algorithms (94.6\% and 95.05\% respectively) reflect how effectively the keystroke dynamics can be used in authenticating a genuine user.

\section{Related work}

Keystroke dynamics has been taken in consideration from a long time and a lot of research has been done on the topic. In 2009, Kevin Killourhy and Roy Maxion have collected the benchmark data set for keystroke dynamics using windows PC and physical keyboard to record the timing vector of a particular password for 51 users. Many researchers have previous worked on this data set to extract the most information. In 2009, Kevin and Roy themselves tried to perform anomaly-detection algorithms to distinguish the imposters from the genuine users. The results demonstrated by the algorithms were not up to the expectations, the ANN algorithm worked the worst.

In late 2009, a research was conducted on dynamic keystroke in  mobile phone to identity users\cite{zahid2009keystroke}. They used fuzzy classifier which resulted in distinguishing 3 different features (key hold time, time difference between pressing and releasing a key and time difference between release and pressing a key). They tried different algorithms like Naive Bayes, Back Propagation Neural Network, K start, Radial Basis Function Network.

Later, a research was done on user's emotional state classification\cite{emotional}. The data set for this research was collected by the research group. The aim of the work was to classify a user's mood into emotional classes like confidence, relaxation, sadness, anger, excitement tiredness and hesitance. Different machine learning classifiers were used and resulted in a max accuracy of 88\%. 

There has been research in age-group classification\cite{age} and checking for fatigue based \cite{fatigue} on typing pattern of a user. There are various other research going on based on keystroke dynamics in security and other user specific classification. The most interesting research ouput is by Yohan Muliono, Hanry Ham and Dion Darmawan. The aim of the work was to classify users based on keystroke dynamics of the users. The group performed different machine learning algorithms like Support vector machine with linear, RBF and Poly kernel, standard deep learning and deep learning algorithm with modified adam optimizer. The group was able to reach a max accuracy of 92.6\% which was the mark to improve.

\section{Data}
\subsection{Data collection details}

The data-set used for the various models was taken from the paper \cite{originalpaper}. It consists of typing samples of 51 users each typing 400 repetitions of the password ".tie5Roanl". The password selected was representative of typical, strong password.  The data was collected in 8 data-collection sessions (50 passwords for each), with at least one day between the sessions, to capture some of the day-to-day variation of each subject's typing. The data is a point in p-dimensional space, where p is the number of features in the timing vectors. So, training data is a cloud of points. 

\subsection{Subject details}

The set of subjects/users consisted of 30 males and 21 females, 8 left-handed and 43 right-handed.The median age group was 31-40, the youngest was 18-20 and oldest was 61-70. The subjects' took between 1.25 and 11 minutes, with the median session taking about 3 minutes.
\subsection{Data-set Features}

The data consists of timing features of users represented as keydown-keydown times, keyup-keydown times, hold times and Enter key times. Total number of timing features extracted were 31, they were stored in seconds(as floating-point numbers).
\begin{figure}[!h]
    \centering
    \includegraphics[width = 0.5\textwidth]{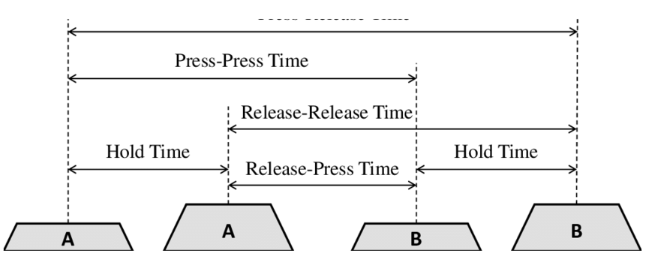}
    \caption{Timing features considered in the data set}
    \label{fig:my_label}
\end{figure}
\subsection{Data pre-processing}

Data was inspected for outliers, missing values and noise values. Ouliers were found in the data set which were removed. The data did not have any missing values or noise and it was found to be normally distributed.

\begin{figure*}[!t]
    \centering
    \includegraphics[width=0.7\textwidth]{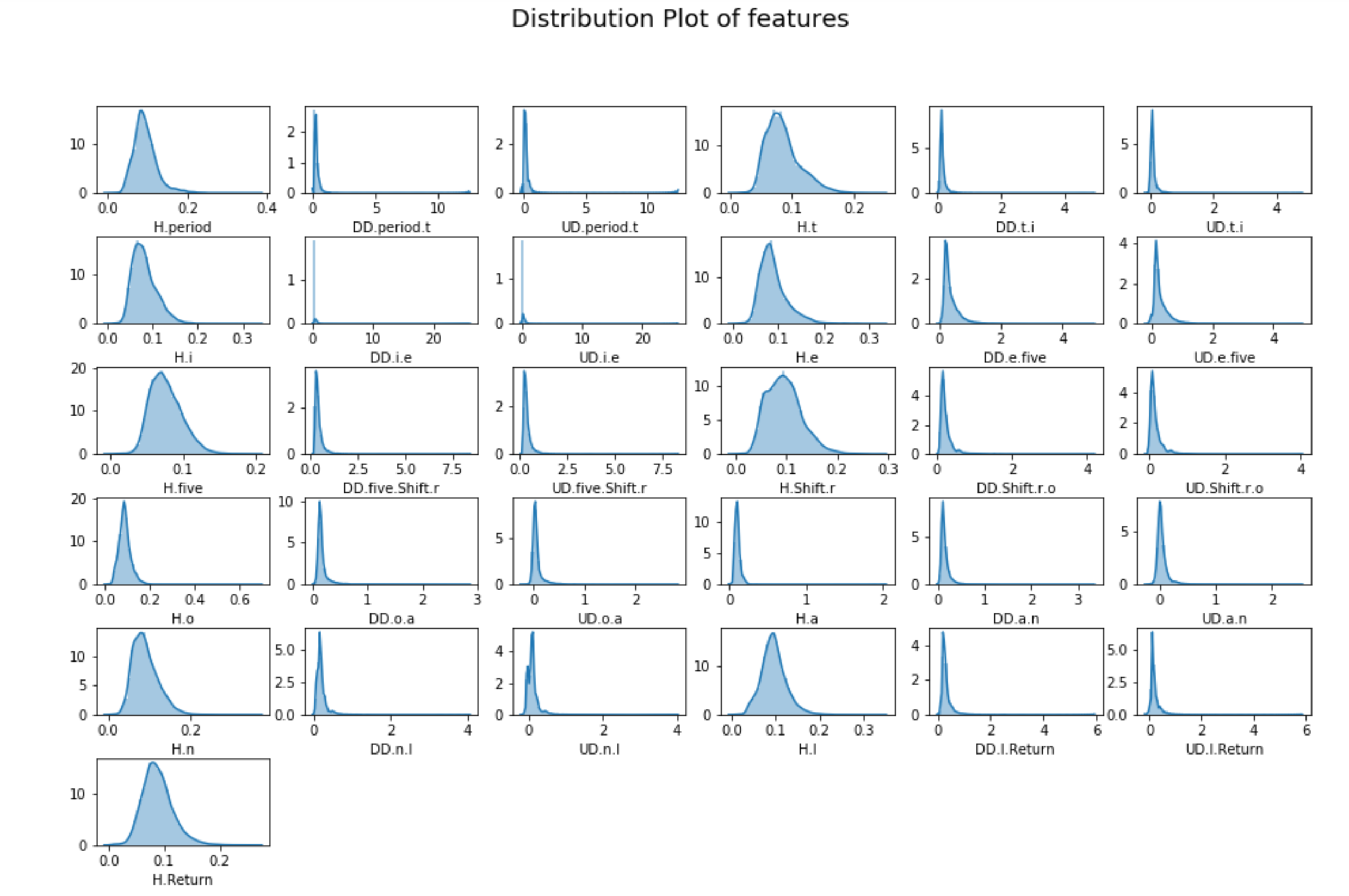}
    \caption{Data Distribution}
\end{figure*}
\begin{figure}[!h]
    \centering
    \includegraphics[width=0.5\textwidth]{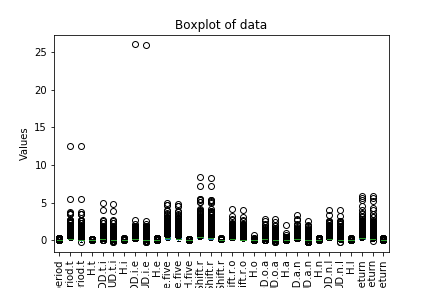}
    \caption{Outlier Detection}

\end{figure}
\section{Methods}
In our data-set we had total of 51 users. Our goal was to implement an anomaly detection system and then extend the same to multi-class classification problem so that we are able to identify user only by using keystroke data.

By using neural networks we have mainly concentrated our efforts in three categories. First, develop a model for multi-class classification problem where we have tried to classify one user among the 51 given use. Second we have tried to detect an anomalous user which the model has never seen before during training using neural network. Third we tried to optimize our models for better accuracy and performance.

For multi class classification our approach is to first try fully connected neural network with two hidden layers. Next tune the model by varying the no of nodes, learning rates and epoch to find optimum hyper parameters. Second we try 1-dimensional convolution layers.  Since during typing of password one key is pressed after the other it made logical sense to try to capture the spatial relation, if there were any, in the features (which are keystroke timings) by using 1D convolution layers. Next tune the model by varying  number of layers, kernel size and channels in each layers learning rate to find optimum hyper parameters. 

In the next section we will try to implement the concept of negative class for anomaly detection. This will just be an experimentation to see if we are able detect an user the model has never seen before and categorize it as an anomalous user. In this approach instead of using all the 51 users for training a classification model we will use only 31 users. The 31 users represent 31 classes. One more class, called the negative class, is created which has random samples from the remaining (51-31) 20 users. Our approach is to see if the model correctly classifies an anomalous user to the negative class.

Finally we improve the model performance by changing the hyper parameters like no of nodes, no of channels, kernel size. We will also be implementing a learning rate scheduler to schedule a learning rate decay when the validation loss plateau.
\section{Experiments}
We implemented all the best performing algorithms from the above cited paper and going further using our learning from NN based object removal\cite{dynamic} implemented neural network, SVM and random forest. We describe the models that we implemented in in this section. 

\subsection{Euclidean}
It is a anamoly detection model. Training data is used to get the centre of the cloud, which is a mean vector. The anomaly score of the test vector is based on its proximity to this mean vector, calculated as the squared Euclidean
distance between the test vector and the mean vector.

\subsection{Manhattan}
The model works similar to the Euclidean model, expect that the distance metric used is manhattan, where the distance between two points is measured along axes at right angles.

\subsection{Manhattan (Scaled)}
As describe in the paper that we referred, anomaly score is calculated as $$\sum_{i=1}^{p} 2^{|x_i - y_i|/a_i} = 1$$ where xi and yi are the i-th features of the test and mean vectors respectively, and ai is the average absolute deviation from the training phase.The score resembles a Manhattan-distance calculation, except each dimension is scaled by ai.

\subsection{Mahalanobis}
In this model, during the training phase, mean vector and co-variance matrix of variables are calculated. The mahalanobis is calculated as 

\[(x-y)^{T}*S^{-1}*(x-y)\]

where S is the covariance matrix

\subsection{Mahalanobis (Normed)}
This model works the same way as Mahalanobis, except that the anomaly score is calculated by “normalizing”
the Mahalanobis distance by dividing the distance by 
\[||x||*||y||\]

\subsection{Z-score}
In this model, the detector calculates the mean and standard deviation of each timing feature in training phase. In the test phase, the detector computes
the absolute z-score of each feature of the test vector. The z-score for the i-th feature is calculated as 
\[|x_i - y_i| / s_i\] where $x_i$ and $y_i$ are the i-th features of the test and mean vectors respectively and $s_i$ is the standard deviation from the training phase. The anomaly score is a count of how many z-scores exceed a threshold. 

\subsection{SVM (one-class)}
In this algorithm, the training data is projected into high dimensional space and linear separator is defined.
The anomaly score is calculated as distance between linear separator and test vector that is projected in the same high dimensional space. The hyperparameter \textit{v} was optimised. 

\subsection{Support Vector Machine (SVM)}
It is a multiclass classification problem, that we are implementing for user identifictaion. The intuition behind the multiclass SVM is that, if our classification rule is 
$$y_i = \textrm{argmax}_{j} \;h(x_{i}) * w_{j} + w_{0,j{'}}$$ 
we should simply make sure that if 
$$y_{i} = j, \textrm{then } h(x_{i}) * w_{j{'}} + w_{0,j{'}} $$
is greater than $$ h(x_{i}) * w_{j{'}} + w_{0,j{'}} $$
for all $$j{'}= j $$
by the largest margin, in the same way that
we make sure that $$h(x_{i}) * w_{j} + w_{0,j} \geq 1$$ in the binary SVM.So we can directly optimize over all of our decision boundaries with constraints that enforce it, and the same objective as before (but now summed over all decision boundaries):
\[min_{w1,...,wL1,w0,1,...,w0,Ly,s1,...,sN} \frac{1}{2}\sum_{j=1}^{L_{y}} {||w_{j}||}+\lambda \sum_{j=1}^{N} {s_{i}}\] 

\begin{figure}[!h]
    \centering
    \includegraphics[width=0.3\textwidth]{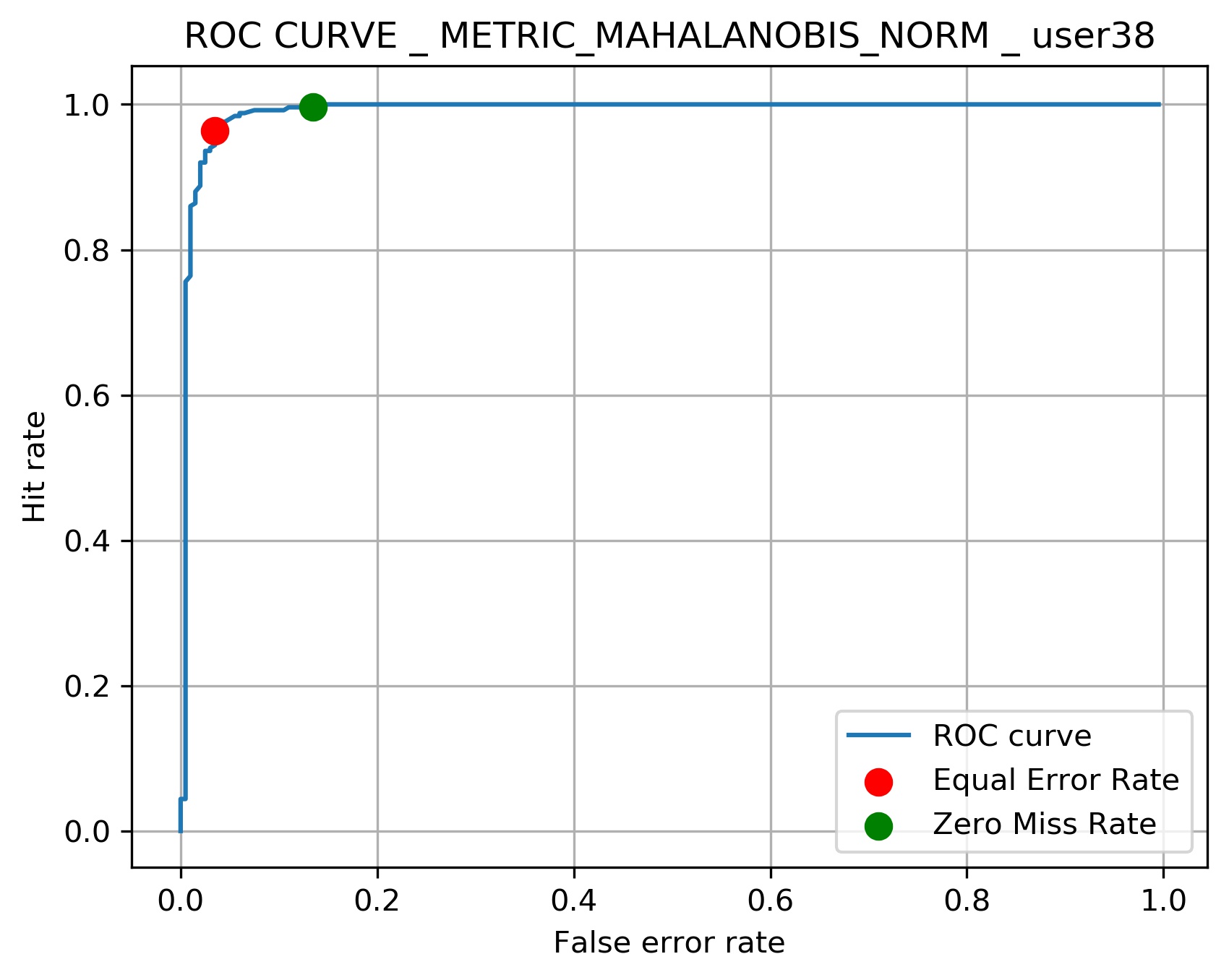}
    \caption{ROC curve of subject 38 using mahalanobis normed}
    \label{fig:my_label}
\end{figure}

The terminologies of ROC Curve are explained below:
\begin{itemize}
  \item Miss rates - Percentage of impostor passwords that are not detected
  \item Hit rate - Frequency with which impostors are detected (i.e., 1 - miss rate)
  \item False-alarm rate - Frequency with which genuine users are mistakenly detected as impostors.
\end{itemize}

\begin{figure*}[!t]
    \centering
    \includegraphics[width=0.8\textwidth]{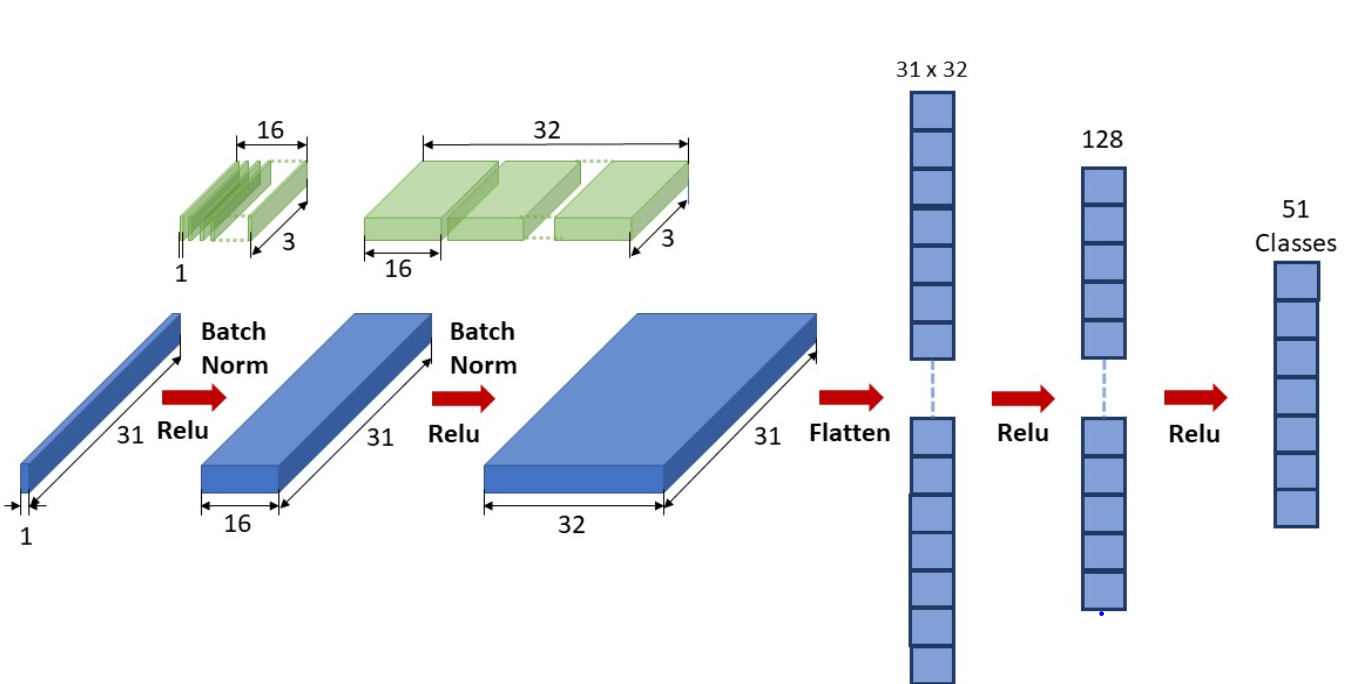}
    \caption{Nerual Network architecture with 1-D convolutional layers}
    \label{fig:my_label}
\end{figure*}

\subsection{Neural Network}
For the calculation of equal-error rate, we chose a threshold so that the detector’s miss and false-alarm rates are equal. Zero-miss false-alarm rate is calculated by choosing the threshold so that the false-alarm rate is minimized under the constraint that the miss rate be zero.

Experimentation of Neural networks were carried out in two phases. First implementation of fully connected network and secondly implementation of 1 dimensional convolutional network.

In fully connected network implementation we have implemented 1 hidden layer and 2 hidden layers configuration for the multi-class classification problem . We also varied the number of nodes in each of the layers to find the most optimum configuration. Based on our test results the best configuration was 2 hidden layers with 80 nodes in first layer and 60 nodes in second layer. The accuracy obtained from the model was 92.2 \% . Learning rate scheduler was used to decrease learning rate with a decay factor of 0.1 when validation loss plateaued.

In 1D convolution we implemented multiple combinations of convolutional layers followed by fully connected layers. After the carrying out hyper parameter tuning the best model configuration was found to be 2 1D convolutional layers back to back. The first layer has 16 channels with kernel size 3. The second convolutional layer has 32 channels with kernel size 3. This is followed by 2 fully connected layers having 992 nodes and 128 nodes. The final layer has 51 nodes corresponding to 51 classes. While training the model a learning rate scheduler was used to decrease the learning rate by a factor of 0.1 when validation loss plateaued. We were able to achieve an accuracy of 94.6\%. This is a more than 2\% accuracy increase than the fully connected configuration. This clearly shows that there is indeed some spatial correlation among the features which has been captured in the 1D convolutional layers.  during trying out various configurations of the network we have also noticed that as we increased the model complexity by increasing the number of convolutional layers or the number of channels in each layer the accuracy drops.
\begin{figure}[!h]
    \centering
    \includegraphics[width=0.4\textwidth]{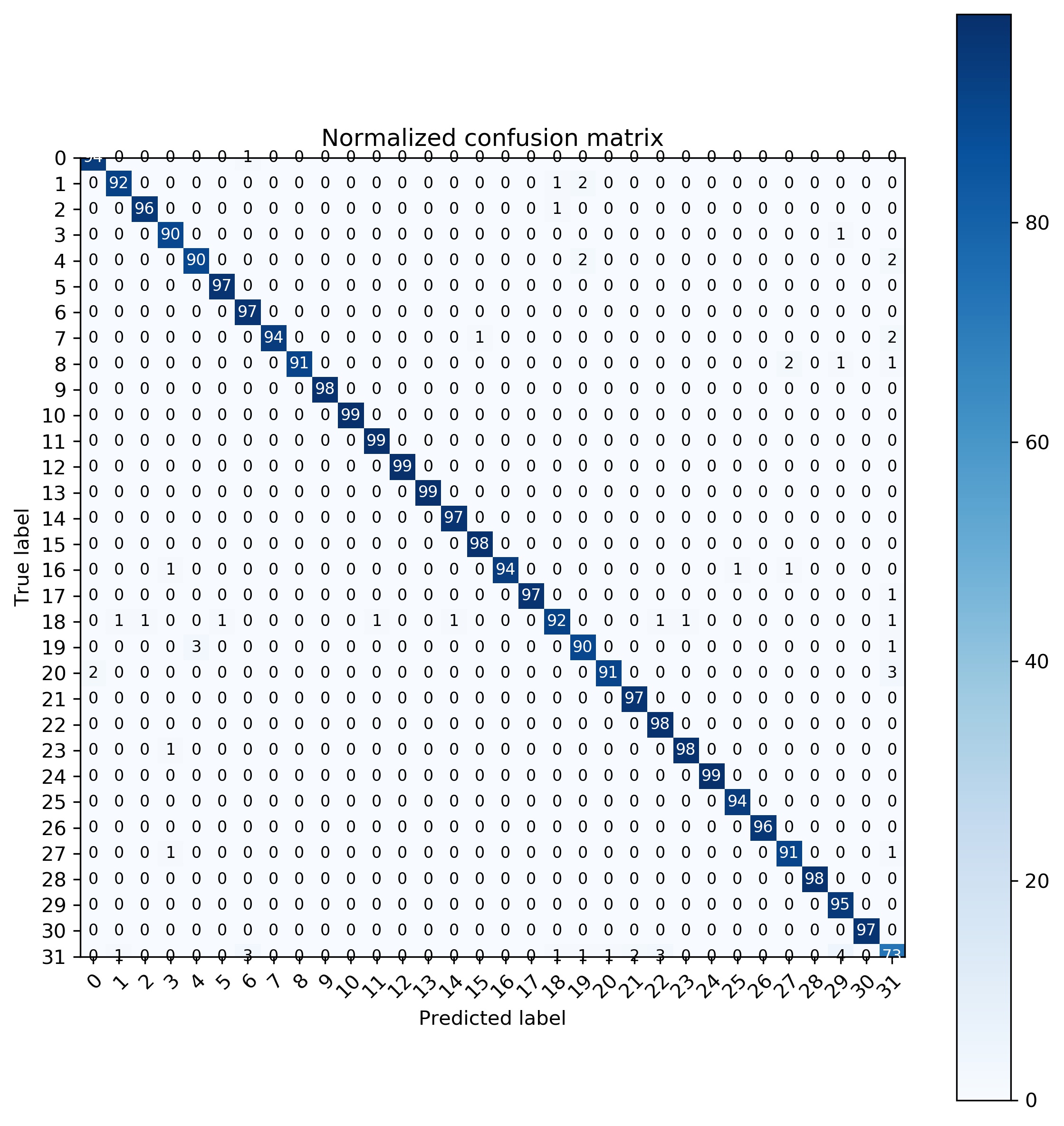}
    \caption{Confusion matrix for ANN with Negative Class}
    \label{fig:my_label}
\end{figure}

Using the above mentioned model configuration we implemented the negative class approach in an attempt to do anomaly detection. In this approach we have taken 31 users of the total 51 users as 31 classes. Next we created one more class, called the Negative Class that is a random combination of all remaining (51-31) 20 classes. Next we train the model using a total of (31+1) 32 classes. Our goal is to see if the trained model is able to classify an anomalous user (which is a data point that doesn't belong to any of the 31 users used to train model) successfully in the negative class. After training the model we were able to get an overall model accuracy of 95.05\% and confusion matrix is shown in figure 6. If we consider the negative class with respect to all other classes, the recall is 80.9\%, precision is 67.22\%, f-score is 0.733.

\subsection{Random Forest Classifier}

For classifying the users into different classes we used Random Forest Classifier. It is an ensemble method of decision trees generated on a randomly split dataset. The dataset was split in the ratio of 75:25 where 75\% was used for training the model and remaining 25\% for testing.Decision trees were constructed for each sample, of which prediction results were obtained.The results with maximum vote was considered as the final prediction. The accuracy of this model on the test data is around 93.66\%
\section{Performance}
The performance matrices of the anomaly-detection algorithms based on distance metrics has been split into Equal Error Rate and Zero-Miss False-Alarm Rate. The results achieved by the algorithms has been listed in the descending order of the average of the parameter. Average Error Rate and standard deviation of Error Rate is shown in Table 1. Average and Deviation of Zero-Miss False-Alarm Rate is shown in Table 2. From the results, we can say that the distance metrics algorithms are not performing well on anomaly detection.
\begin{table}[ht]
\centering
\caption{Table of Average and Standard Deviation (S.D.) of Equal Error Rate (EER).}
\begin{tabular}[t]{lcc}
\toprule
Detector &Average EER&S.D. EER\\
\midrule
Euclidean & 0.265 & 0.079\\
Manhattan & 0.206 & 0.077\\
Mahalanobis & 0.193 & 0.101\\
Mahalanobis (Normed) & 0.166 & 0.092\\
Manhattan (Scaled) & 0.141 & 0.068\\
Z-Score & 0.135 & 0.065\\
\bottomrule
\end{tabular}
\end{table}

\begin{table}[ht]
\centering
\caption{Table of Average and Standard Deviation (S.D.) of Zero-Miss False-Alarm Rate (ZFR).}
\begin{tabular}[t]{lcc}
\toprule
Detector &Average ZFR&S.D. ZFR\\
\midrule
Z-Score & 0.535 & 0.275\\
Manhattan (Scaled) & 0.540 & 0.271\\
Mahalanobis & 0.645 & 0.299\\
Mahalanobis (Normed) & 0.666 & 0.301\\
Manhattan & 0.711 & 0.224\\
Euclidean & 0.753 & 0.212\\
\bottomrule
\end{tabular}
\end{table}

In table 3, we can see the performance of our machine learning classifier algorithms. From the table, we can say Artificial Neural Network with negative class classification performs the best with an accuracy of 95.05\%.
\begin{table}[ht]
\centering
\caption{Table of Accuracy of Machine Learning Models.}
\begin{tabular}[t]{lcc}
\toprule
Multi-Class Classification &Accuracy \%\\
\midrule
ANN (With Negative Class) & 95.05 \\
ANN (1-D Conv.) & 94.6\\
Random Forest & 93.66\\
ANN (Fully Connected) & 92.2\\
Support Vector Machine & 75.6\\
\bottomrule
\end{tabular}
\end{table}

\section{Conclusion}

The results presented here clearly shows that based on the feature vector created from keystroke dynamics it is possible to classify users. Also, results obtained from training neural networks of different configurations we can see that there are some spatial correlation in the feature vector. However the data collected was based on one password that was typed by 51 users. An extension of this work could be to analyse the keystroke dynamics of multiple users over some commonly occurring words in the English language tokened as Flag Words. Once a user starts using the keyboard the keystroke dynamics can be monitored online and in the event any flag word in typed in, the data can be used to verify the authenticity of the user. This feature can be used in smartphones, desktop, laptops or any other systems having a keyboard a an extra layer of security of user identification. As a future work, emotion detection based on keystrokes tactile sensing can be added as extra layer of security, if highly sensitive soft-bodied tactile\cite{soft-body,tactile} sensors are used.

{\small
\bibliographystyle{ieee}
\bibliography{references}

\begin{thebibliography}{10}\itemsep=-1pt

\bibitem{ali2017keystroke}
M.~L. Ali, J.~V. Monaco, C.~C. Tappert, and M.~Qiu.
\newblock Keystroke biometric systems for user authentication.
\newblock {\em Journal of Signal Processing Systems}, 86(2-3):175--190, 2017.

\bibitem{iris}
A.~Bansal, R.~Agarwal, and R.~Sharma.
\newblock Svm based gender classification using iris images.
\newblock In {\em 2012 Fourth International Conference on Computational
  Intelligence and Communication Networks}, pages 425--429. IEEE, 2012.

\bibitem{emotional}
C.~Epp, M.~Lippold, and R.~L. Mandryk.
\newblock Identifying emotional states using keystroke dynamics.
\newblock In {\em Proceedings of the sigchi conference on human factors in
  computing systems}, pages 715--724. ACM, 2011.

\bibitem{fatigue}
J.~Hayes.
\newblock Identifying fatigue through keystroke dynamics.
\newblock {\em The UNSW Canberra at ADFA Journal of Undergraduate Engineering
  Research}, 9(2), 2018.

\bibitem{idrus2014soft}
S.~Z.~S. Idrus, E.~Cherrier, C.~Rosenberger, and P.~Bours.
\newblock Soft biometrics for keystroke dynamics: Profiling individuals while
  typing passwords.
\newblock {\em Computers \& Security}, 45:147--155, 2014.

\bibitem{originalpaper}
K.~S. Killourhy and R.~A. Maxion.
\newblock Comparing anomaly-detection algorithms for keystroke dynamics.
\newblock In {\em 2009 IEEE/IFIP International Conference on Dependable Systems
  \& Networks}, pages 125--134. IEEE, 2009.

\bibitem{muliono2018keystroke}
Y.~Muliono, H.~Ham, and D.~Darmawan.
\newblock Keystroke dynamic classification using machine learning for password
  authorization.
\newblock {\em Procedia Computer Science}, 135:564--569, 2018.

\bibitem{patil2016keystroke}
R.~A. Patil and A.~L. Renke.
\newblock Keystroke dynamics for user authentication and identification by
  using typing rhythm.
\newblock {\em International Journal of Computer Applications}, 144(9):1, 2016.

\bibitem{tactile}
G.~Ponraj, A.~V. Prituja, C.~Li, A.~Bamotra, Z.~Guoniu, S.~K. Kirthika, N.~V.
  Thakor, A.~B. Soares, and H.~Ren.
\newblock Active contact enhancements with stretchable soft layers and
  piezoresistive tactile array for robotic grippers.
\newblock In {\em 2019 IEEE 15th International Conference on Automation Science
  and Engineering (CASE)}, pages 1808--1813, 2019.

\bibitem{revett2007machine}
K.~Revett, F.~Gorunescu, M.~Gorunescu, M.~Ene, P.~S.~T. Magalh{\~a}es, and
  H.~D.~d. Santos.
\newblock A machine learning approach to keystroke dynamics based user
  authentication.
\newblock {\em International Journal of Electronic Security and Digital
  Forensics}, 1(1):55--70, 2007.

\bibitem{age}
S.~Roy, U.~Roy, and D.~Sinha.
\newblock Protection of kids from internet threats: A machine learning approach
  for classification of age-group based on typing pattern.
\newblock In {\em Proceedings of the International MultiConference of Engineers
  and Computer Scientists}, volume~1, 2018.

\bibitem{dynamic}
P.~K. Uppala, A.~Bamotra, and R.~Kolamuri.
\newblock Dynamic object removal for effective slam, 2023.

\bibitem{security}
R.~Von~Solms and J.~Van~Niekerk.
\newblock From information security to cyber security.
\newblock {\em computers \& security}, 38:97--102, 2013.

\bibitem{soft-body}
P.~Walia, A.~Bamotra, A.~Prituja, and H.~Ren.
\newblock Design and fabrication of soft-bodied 3-d tactile sensors with
  magnetometers.
\newblock In {\em 2018 IEEE International Conference on Information and
  Automation (ICIA)}, pages 1284--1289, 2018.

\bibitem{zahid2009keystroke}
S.~Zahid, M.~Shahzad, S.~A. Khayam, and M.~Farooq.
\newblock Keystroke-based user identification on smart phones.
\newblock In {\em International Workshop on Recent advances in intrusion
  detection}, pages 224--243. Springer, 2009.

\end{thebibliography}
}

\end{document}